\documentclass[sigconf]{acmart}
\usepackage{multirow}
\DeclareMathOperator{\E}{\mathbb{E}}
\newcommand{\norm}[1]{\left\lVert#1\right\rVert}
\usepackage[ruled,vlined,linesnumbered]{algorithm2e}
\newcommand{\Lagr}{\mathcal{L}}
\usepackage{subfigure}
\AtBeginDocument{%
  \providecommand\BibTeX{{%
    \normalfont B\kern-0.5em{\scshape i\kern-0.25em b}\kern-0.8em\TeX}}}

\setcopyright{none}

\copyrightyear{2021}
\acmYear{2021}
\setcopyright{acmcopyright}\acmConference[MMAsia '20]{ACM Multimedia Asia}{March 7--9, 2021}{Virtual Event, Singapore}
\acmBooktitle{ACM Multimedia Asia (MMAsia '20), March 7--9, 2021, Virtual Event, Singapore}
\acmPrice{15.00}
\acmDOI{10.1145/3444685.3446302}
\acmISBN{978-1-4503-8308-0/21/03}

\begin{document}

\title{C3VQG: Category Consistent Cyclic Visual Question Generation}

\author{Shagun Uppal$^{1*}$, Anish Madan$^{1*}$, Sarthak Bhagat$^{1*}$, Yi Yu$^{2}$, Rajiv Ratn Shah$^{1}$}\thanks{$^*$Equal contribution. Ordered Randomly}
\affiliation{%
\institution{$^1$IIIT-Delhi, India; $^2$NII, Japan}
}
\email{{shagun16088, anish16223, sarthak16189, rajivratn}@iiitd.ac.in, yiyu@nii.ac.jp}


\renewcommand{\shortauthors}{Uppal et al.}

\begin{abstract}
Visual Question Generation (VQG) is the task of generating natural questions based on an image. Popular methods in the past have explored image-to-sequence architectures trained with maximum likelihood which have demonstrated meaningful generated questions given an image and its associated ground-truth answer. VQG becomes more challenging if the image contains rich contextual information describing its different semantic categories. In this paper, we try to exploit the different visual cues and concepts in an image to generate questions using a variational autoencoder (VAE) without ground-truth answers. Our approach solves two major shortcomings of existing VQG systems: (i) minimize the level of supervision and (ii) replace generic questions with category relevant generations. Most importantly, by eliminating expensive answer annotations, the required supervision is weakened. Using different categories enables us to exploit different concepts as the inference requires only the image and the category. Mutual information is maximized between the image, question, and answer category in the latent space of our VAE. A novel category consistent cyclic loss is proposed to enable the model to generate consistent predictions with respect to the answer category, reducing redundancies and irregularities. Additionally, we also impose supplementary constraints on the latent space of our generative model to provide structure based on categories and enhance generalization by encapsulating decorrelated features within each dimension. Through extensive experiments, the proposed model, ${\tt C3VQG}$ outperforms state-of-the-art VQG methods with weak supervision.
\end{abstract}

\ccsdesc[500]{Computing methodologies~Machine learning}
\ccsdesc[500]{Computing methodologies~Computer vision}
\ccsdesc[300]{Computing methodologies~Natural language processing}

\keywords{visual question generation, cycle consistency, multimodal}


\maketitle

\section{Introduction}

Visual understanding by intelligent systems is an interesting problem in the Computer Vision and Multimedia community, further accelerated by the advent of Deep Learning \cite{Uppal2020EmergingTO,Shah2017MultimodalAO}. 
Translating visual understanding into language helps us evaluate the "comprehension capability" of the system. Tasks like Visual Question Answering (VQA) \cite{Agrawal2015VQAVQ,malinowski2014multiworld,zhu2015visual7w}, Visual Question Generation (VQG) \cite{Mostafazadeh2016GeneratingNQ}, Video Captioning \cite{Chen2019DeepLF}, and Text-Conditioned Image Generation \cite{Nasir2019Text2FaceGANFG,Pai2019UserIB} help us benchmark it. Such tasks require us to learn multimodal VisLang representations. VQG is a more open-ended and creative task than VQA, in the sense that asking semantically coherent and visually relevant questions requires a system to recognize various concepts present in an image. Contrary to this, in VQA the model tries to infer specific cues from the given inputs in order to answer the reference questions. 

\begin{figure}
    \centering
    \includegraphics[scale=0.33]{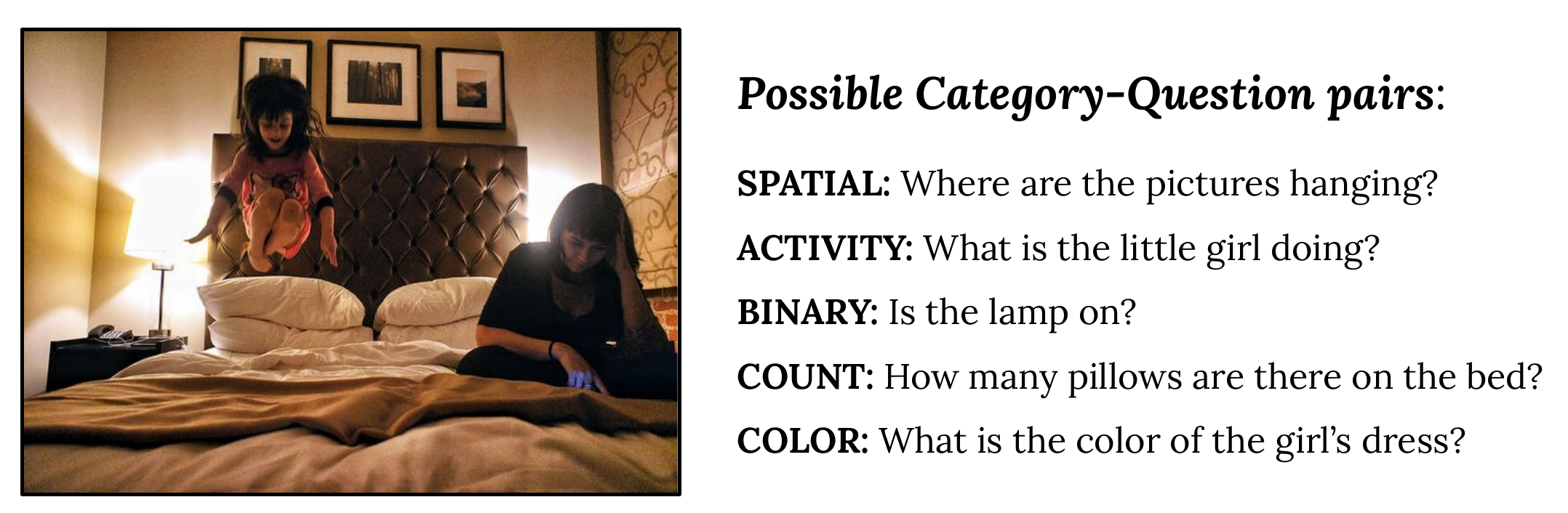}
    \caption{An example image showing various natural questions possible as per mentioned categories. The categories are not too specific so as to overly-constrain the network but broad enough to encourage discovery of novel concepts. }
    \label{fig:intro}
\end{figure}

Figure~\ref{fig:intro} illustrates some abstract concepts and the various semantics that are captured via broad categories considered for question generation. Each category is distinctive enough to be exclusive from others and at the same time, covers a broad range of possibilities for question generation, when an image is conditioned over it. 

\par

Developing a solution for VQG requires one to model novel conceptual discoveries about language and visual representations which pose certain challenges: (1) There are various visual concepts in the images, (2) Questions generated need to be relevant to the image, (3) The generated question to image relation is many-to-one since multiple questions are possible for an image, and (4) Avoiding questions which invoke generic answers like "yes" or "I do not know". For \textit{e.g.}, in Figure~\ref{fig:intro}, we can observe the little girl jumping, the mother trying to read something, the image is of a hotel room, there are photos hanging on top of the bed, \textit{etc}. The questions in the figure satisfy the above criteria.

For attaining human-level understanding of multimodal real-world data, system designs should be created in order to overcome such challenges. This is the reason the task of VQG has also been referred as a realization of the Visual Curiosity \cite{Yang2018VisualCL} of a system. \par

Previous works \cite{Li2017VisualQG,liu2018ivqa,xu2018dual,Krishna2019InformationMV} often use answers along with the image to generate relevant questions. While these approaches ask questions relevant to the image (due to the answer being provided), it tends to overfit to the answer provided and does not leave room for creatively generating questions.
This requires the dataset to be annotated with answers as well as questions which is an expensive and tedious operation. \par

While current works rely heavily on the availability of question-answer pairs for their method, we propose using categories instead of answers. This incorporates a weaker form of supervision, which is easy to obtain and helps constrain the problem to enforce relevance of generated questions to an image as opposed to those without constraints \cite{Mostafazadeh2016GeneratingNQ,Zhang2016AutomaticGO,Jain2017CreativityGD}. We propose a category-specific generative modelling framework which makes multiple relevant category-specific question generations per image possible. 

The following are the main contributions of the paper: 
\begin{itemize}

     \item We weaken the amount of supervision on the model by removing the need of ground-truth answers during training. 
    
    \item We adopt a variational autoencoder \cite{kingma2013auto} framework to generate questions using a combined latent space for image and category and maximize mutual information between them.
   
    \item We introduce additional constraints to enforce answer category consistency using a cyclic training procedure with sequential training in two disjoint steps. 
    
    \item We enforce center loss on the generative latent space to ensure clustering with respect to answer category labels, making generations more category-specific and robust. 
    
    \item We also introduce a hyper-prior on learning the inverse variance of variational latent prior to capture intrinsically independent visual features within the combined latent space. 
\end{itemize}

Our contributions ensure diverse (see Section \ref{quan_results}) and relevant (see Section \ref{qual_results}) question generations given an image and category. We evaluate our result with other recent approaches which do not use answers for question generation as well as which require them. \par


\section{Related Works}
\label{related_Works}

In this section, we discuss relevant literature that motivates key components of the ${\tt C3VQG}$ approach. 
In Section \ref{vqg}, we focus on various approaches that emphasised on the task of question generation from visual inputs. This is followed by Section \ref{latent_structure}, where we describe appropriate studies that have remodelled their latent representations for the escalation of downstream task performance. 

\subsection{Visual Question Generation (VQG)}
\label{vqg}

VQG is the task of developing visual understanding from images using cues from ground-truth answers and/or answer categories in order to generate relevant question. Various works focusing on this aspect have been deeply inspired by taking into consideration the multimodal context of natural language along with visual understanding of the input.

Mostafazadeh \textit{et al.} \cite{Mostafazadeh2017ImageGroundedCM} suggested relevant question-response generation, given an image along with relevant conversational dialogues. Using dialogues, they drew broad context about the conversation from the input image. Mostafazadeh \textit{et al.} \cite{Mostafazadeh2016GeneratingNQ} focused on a different paradigm of VQG to generate engaging high-level commonsense reasoning questions about the event highlighted in the visual input. The approach shifted its focus from objects constituting the image to visual understanding of systems. 

Yang \textit{et al.} \cite{Yang2015NeuralST} simultaneously learned \emph{VQG} and \emph{VQA} models to understand the semantics and entities present in the input image. Such an approach examined and trained the learning model on both the aspects of language and vision, thereby, challenging its interpretability over multimodal signals. Li \textit{et al.} \cite{Li2017VisualQG} had a similar approach of training \emph{VQA} and \emph{VQG} networks parallely, hence, introducing an Invertible QA-network. Such a model took advantage of the QA dependencies while training, then took a question/answer as an input, outputting its counterpart for evaluation. Works like \cite{Wang2017AJM}, synchronized both the tasks to learn co-operatively but restricted their abilities to explore non-trivial aspects of generation.

Zhang \textit{et al.} \cite{Zhang2016AutomaticGO} talked about automating \emph{VQG} not only with high correctness but with a high diversity in the type of questions generated. They took an image and its caption as the input. The question type along with the input image, caption and their correlation output were processed to output relevant questions. Similarly, Jain \textit{et al.} \cite{Jain2017CreativityGD} worked on generating multiple questions given an image using generative modelling. Here, they used a VAE with a set of LSTM networks in order to generate a diverse set of questions.

While prior work in VQG has spanned a wide variety of training strategies for meaningful question generation, our approach ${\tt C3VQG}$ is unique in the sense that it utilizes a mutual information maximization technique with weak supervision. On top of it, it learns a well-structured latent space with a non-standard Gaussian prior and category-wise clustering. 

\subsection{Structured Latent Space Constraints}
\label{latent_structure}

\subsubsection{Center Loss for Learning Discriminative Latent Features}
\label{center_loss_related}

Center loss \cite{Wen2016ADF} for enforcing well-clustered latent spaces have been studied extensively in the past specifically for biometric applications \cite{Wen2016ADF, Wen2018ACS, Kazemi2018AttributeCenteredLF}. This metric-learning training strategy works on the principle of differentiating inter-class features and penalizing embedding distances from their respective class centers. 

Wen \textit{et al.} \cite{Wen2018ACS} utilized center loss for the biometric task of facial recognition. The introduction of weight sharing between softmax and the center loss reduces the computational complexity. While, the employment of an entire embedding space as the center rather than the conventionally used single point representation takes into account the intra-class variations as well. Kazemi \textit{et al.} \cite{Kazemi2018AttributeCenteredLF} also proposed a novel attribute-centered loss in order to train a Deep Coupled CNN for sketch-to-photo matching using facial features. 

He \textit{et al.} \cite{He2018TripletCenterLF} proposed a triplet-center loss that aims at further improving the differentiating power of features by not only minimizing the distance of encoding from their class centers but also by maximizing it for the class centers belonging to other classes. 
Ghosh and Davis \cite{Ghosh2018UnderstandingCL} highlighted the impact of introduction of center loss besides the cross entropy loss in CNNs for image retrieval problems, involving very few samples belonging to each class.

\subsubsection{Hyper-prior on Latent Spaces}
\label{bayes_related}

Various approaches have intended to capture completely decorrelated factors of variations in the data using diverse training strategies such as generative models to learn low-dimensional subspaces \cite{Klys2018LearningLS} or imposing a soft orthogonality constraint on latent chunks \cite{Shukla2019ProductOO}. One such effective approach is to vary the prior on the generative latent space in such a way that it intrinsically enforces independence of captured features.

Kim \textit{et al.} \cite{Kim2019BayesFactorVAEHB} introduced a class of hierarchical Bayesian models with certain hyper-priors on the variances of the Gaussian distribution priors in a VAE. The fact that this ensures that each captured latent feature has a different prior distribution ensures that each of them are intrinsically independent and guarantees encapsulation of admissible as well as nuisance factors simultaneously. 
Ansari and Soh \cite{Ansari2018HyperpriorIU} also focused on capturing disentangled factors of variations in an unsupervised manner by utilizing Inverse-Wishart (IW) as the prior on the latent space of the generative model. By tweaking the IW parameter, various features in a set of diverse datasets could be captured simultaneously.
Bhagat \textit{et al.} \cite{Bhagat2020DisentanglingRU} utilized Gaussian processes (GP) with varying correlation structure in VAEs for the task of video sequence disentangling. In general, structured latent spaces has aided downstream task performance in diverse fields such as image captioning \cite{10.1145/3295748} and language inferences \cite{li2019consistency,uppal-etal-2020-two}. 

To the best of our knowledge, center loss for latent clustering on the latent space for capturing independent factors of variation has never been deployed in a multimodal setting. We take motivation from several works that have utilized these techniques to formulate a structured latent representation in order to wield superior performance on downstream tasks.

\section{Proposed Approach}
\label{sec:proposed_aproach}

We introduce \textbf{${\tt C3VQG}$} \footnote{We provide the entire algorithm for training in the supplementary material.}, a question generation architecture which only requires <images, questions, categories> for training, and <images, category> for inference. We propose a cyclic training approach that enforces consistency in answer categories via a two-step framework. For this, we introduce a VAE-setting which maximizes mutual information between the question generated, image and category. 

The training flow \footnote{A similar illustration for the inference framework is provided in the supplementary.} is illustrated in Figure \ref{fig:train_ours}.
We divide the training architecture into two disjoint steps. While the first step ensures encapsulation of image and category information within the latent encoding, the second step establishes compatibility in predicted categories from the generated question with that of the ground-truth categories. 
We enforce the latent space to capture independent features in the image in a structured manner with an additional hyper-prior (refer Section \ref{latent_hyper_prior}) and a center loss based constraint (refer Section \ref{center_loss_clustering}). While the former maintains a high diversity across generated questions, the latter helps in maintaining relevance between image, answer-category and the generated question. 

\subsection{Problem Formulation}

For accomplishing this task of generating meaningful questions from multimodal sources of data in the form of images and answer categories, we have training data in the form of images and corresponding question from different answer categories. We denote all unique images by the set $I_{D}$, set of all unique answer categories by $C_{D}$, and set of all unique ground-truth questions by $Q_{D}$, where length of the sets are given by $n_{I}$, $n_{c}$, and $n_{q}$ respectively.
We define our training dataset as a collection of  $n$ 3-tuples, $dset = \{<i_{1},q_{1},C_{1}>, ..., <i_{n},q_{n},C{n}>\}$. For the $k^{th}$ sample in our dataset, we have image $i_{k} \in I_{D}, q_{k} \in Q_{D}, C_{k} \in C_{D}$,  as $C_{D} = \{C_{1}, C_{2} ... C_{n_{c}}\}$.

We denote the predicted question as $\hat{q}_{k,C}$, where $k$ denotes the sample for which the question is predicted and $C$ denotes the category ($C \in C_{D}$), as we generate $n_c$ questions for every sample in our training set. We also denote our latent space by $z$, and the dimensions of the combined latent space by $d$.

\subsection{Information Maximisation \textit{VQG}}

We consider the case of a single image $i$, its corresponding category $C$ and the question we want to generate $q$.
We define our initial model (referred as Step I in Section \ref{C3VQG}) by defining $p(q|i,C)$ which we get by maximizing a linear combination of mutual information $I(i,q)$ and $I(C,q)$. To avoid optimizing the gradient in discrete steps (in order to get low bias and variance of the gradient estimator), we try to learn a mapping $p_\phi(z|i,C)$ from the image and category to a continuous latent space which we refer to as $z$. The mapping is parameterized by $\phi$ which is learned via optimization of the following objective:

\begin{align}\label{eqn:mu_inf}
    \max_\phi \quad I(q,z|i,C) + \lambda_1 I(i,z) + \lambda_2 I(C,z) \\
    s.t \quad z \sim p_\phi(z|i,C) \hspace{2mm} \text{and} \hspace{2mm}
    q \sim p_\phi(q|z)
\end{align}

where $\lambda_{1}$ and $\lambda_{2}$ are the weights for the mutual information terms.
The mutual information in Equation \ref{eqn:mu_inf} is intractable as we do not know the true values of the posteriors $p(z|i)$ and $p(z|C)$. So we instead try to minimize its variational lower bound (ELBO). More details on the derivation of the final objective can be found in the supplementary section.
Hence, we can optimize the variational lower bound by maximizing the image and category reconstruction whilst also maximizing the MLE of question generation.

\subsection{Category Consistent Cyclic VQG (C3VQG)}
\label{C3VQG}

\begin{figure*}
        \centering
        \includegraphics[scale=0.47]{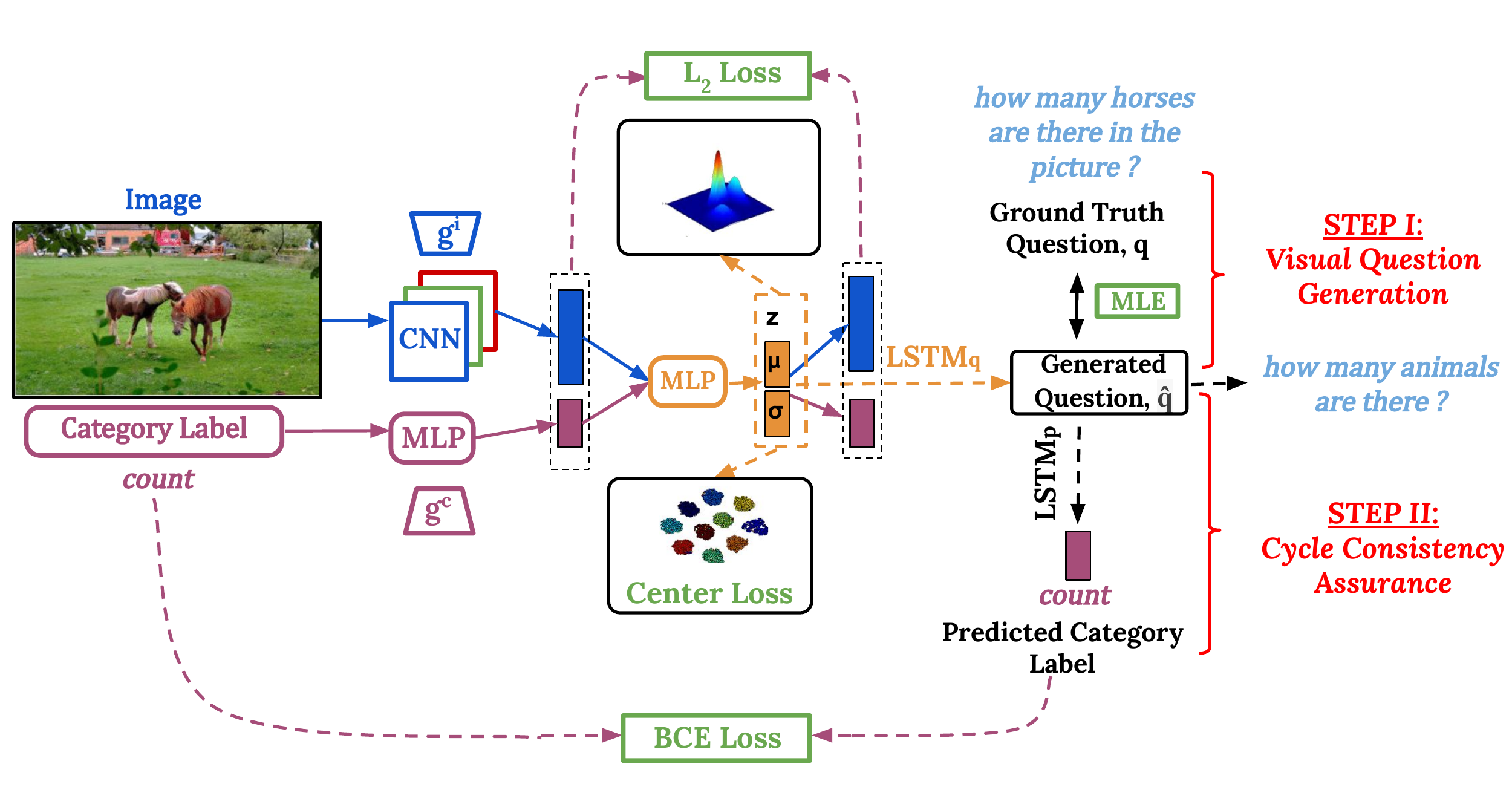}
        \caption{${\tt C3VQG}$ Training Framework}
        \label{fig:train_ours}
\end{figure*}

We build a cyclic approach for VQG to analyze the robustness of the model in terms of its predictions and the diversity of generated questions. For this, we divide our approach into two parts. The first step homogenizes the latent representations obtained from the answer categories and the one obtained from images to form a combined latent space. While, the next step penalises the difference in ground-truth answer categories from the ones predicted from the generated question, enforcing congruence between them.

\paragraph{Step 1: Visual Question Generation.}

Using two separate encoders $g^{i}$ and $g^{c}$, we generate latent encoding $h_{k}^{i}$ and $h_{k}^{c}$ for the image $i_{k}$ and category label $C_{k}$ respectively.

\begin{align}
\label{encodings}
    h_{k}^{i} = g^{i}(i_{k}) \hspace{2mm} \text{and} \hspace{2mm}
    h_{k}^{c} = g^{c}(C_{k})
\end{align}

These latent encodings are passed onto an MLP after concatenation to generate another latent representation that has a Gaussian prior associated with it. The latent representation $z \in \mathbb{R}^d$ forms the backbone for question generation, and is given by Equation \ref{latent_eq}.

\begin{align}
\label{latent_eq}
    z_{k} = \mathbf{W_{MLP}}^\intercal(h_{k}^{i} \oplus h_{k}^{c})
\end{align}

where $\mathbf{W_{MLP}}$ depicts the weights of the MLP and $\oplus$ depicts the concatenation operator for two input vectors. 
The concatenation of the two encodings aggregates the category information along with the visual cues for question generation. This latent encoding should intrinsically contain all the relevant information for the generation of the question. Therefore, it is passed through an LSTM that outputs the question related to the images on the lines of the answer category. 

\begin{equation}
\label{question_generation}
    \hat{q}_{k,C_k} = LSTM_{q}(z_{k})
\end{equation}

Therefore, we capitalise on the ground-truth question $q_k$ for the image to impose an MLE loss on the generated question $\hat{q}_{k,C_k}$.

\begin{equation}
\label{questionloss}
    \Lagr_{Q} = \norm{\hat{q}_{k,C_k} - q_{k}}_{2}^{2}
\end{equation}

In order to ensure abbreviation of visual features as well as category information into the $z$-space, we pass it through two separate prediction networks, $p^{i}$ and $p^{c}$ respectively. These prediction networks are trained to reconstruct the original image and category encodings.

\begin{equation}
\label{imgandcategoryloss}
    \Lagr_{I} = \norm{p^{i}(z_{k}) - h_{k}^{i}}_{2}^{2} \hspace{2mm} \text{and} \hspace{2mm}
    \Lagr_{C} = \norm{p^{c}(z_{k}) - h_{k}^{c}}_{2}^{2}
\end{equation}

\paragraph{Step 2: Generation Consistency Assurance.}

In order to substantiate the consistency of the answer category of the generated question with the given category, we pass the generated question $\hat{q}_{k,C_k}$ through a temporal classifier $LSTM_{p}$ that tries to predict the answer category for the generated question.

\begin{equation}
\label{cat_pred}
    C^{pred}_{k} = LSTM_{p}\Big(\hat{q}_{k,C_k}\Big)
\end{equation}

Later, we impose a cross entropy loss between the predicted and actual answer category in order to penalise any irregularities within the previous step.

\begin{equation}
\label{consistencyloss}
    \Lagr_{cons} = - C_{k} \log{C^{pred}_{k}}
\end{equation}

\subsection{Latent Space Clustering}
\label{center_loss_clustering}

To ensure that our model is able to accurately predict answer categories from the latent encodings, we intend to promote well-clustered latent spaces. For this, we add structure to the latent space by imposing a constraint in the form of center loss \cite{Wen2016ADF} that aggregates the latent space into a fixed number of clusters, equal to the number of answer categories in the dataset. 

The center loss helps distinguish inter-category latent features by enforcing clustering in the following way:

\begin{equation}
    \Lagr_{center} = \norm{z_{k} - c_{k}}_{2}^{2}, 
    \label{center_loss}
\end{equation}

\noindent where, $c_{k} \in \mathbb{R}^{d}$ depicts the class center for all such datapoints $z_{k}$ (where, $k \in [1, n]$) with label $C_{k}$. These centers are obtained by averaging the features of the corresponding classes, updated based on mini-batches instead of the entire training data due to computational time constraints. Additionally, the update of these centers are scaled by a constant ($< 1$) to avoid sudden fluctuations. 
The structured latent representation that is obtained as a result of applying this constraint ensures escalation of distances in the latent space between samples belonging to different classes, that in turn leads to enhanced downstream task performance. 

\subsection{Modified Hyper-prior on the Latent Space}
\label{latent_hyper_prior}

We also take motivation from one of the models proposed by Kim \textit{et al.} \cite{Kim2019BayesFactorVAEHB} that introduced a modified prior on the latent space explicitly ensuring each dimension to capture independent features. We do this by replacing the sub-optimal Gaussian normal prior on the $z$-space by a long-tail distribution. We introduce a learnable hyper-prior on inverse variance of the Gaussian latent prior while keeping the distribution as zero mean. We also employ a supplementary regularization term that ensures sufficient nuisance dimensions.

For this, we intend to learn the inverse variance $\alpha_{j}$ for each dimension $j$ of the $d$-dimensional latent space. The latent space prior can then be represented as Equation \ref{bayes_1}.

\begin{equation}
    p(z_{k} | \alpha) = \prod_{j=1}^{d} p(z_{k,j} | \alpha_j) = \prod_{j=1}^{d} \mathcal{N}(z_{k,j}; 0, \alpha_{j}^{-1})
\label{bayes_1}
\end{equation}

\noindent Here, $z_{k,j}$ represents the $j^{th}$ dimension of the vector $z_{k} \in \mathbb{R}^{d}$.

The modified KL-divergence and additional regularization term is of the form given by Equation \ref{bayes_loss}.

\begin{equation}
\begin{split}
\label{bayes_loss}
    \Lagr_{bayes} = \sum_{j=1}^{d} \E_{pd(x^{cc}_{k})}\big[ KL(f(z_{k,j}|x^{cc}_{k,j}) || \mathcal{N}(z_{k,j}; 0, \alpha_{j}^{-1})) \big] \\ + \lambda_{reg} \sum_{j=1}^{d} (\alpha_{j}^{-1} - 1)^{2}
\end{split}
\end{equation}

\noindent Here, $x_{k}^{cc}$ is the concatenated latent encoding of the image and category encoding, i.e., $h_{k}^{i} \oplus h_{k}^{c}$, $x_{k,j}^{cc}$ depicting its $j^{th}$ dimension, $z$ is the latent encoding with variational prior, and $f$ is the mapping function (i.e., $f: x^{cc} \rightarrow z$). The expectation is taken over the entire probability distribution ($pd$) of $x^{cc}_{k}$ $\forall k \in [1, n]$.
In Equation \ref{bayes_loss}, $\lambda_{reg}$ is the weight for the regularization loss that promotes sparsity and increases generalization capacity of the model. 






\section{Evaluation}
\label{evaluation}

We evaluate the performance of our approach ${\tt C3VQG}$ \footnote{Code available at \href{https://github.com/sarthak268/C3VQG-official}{https://github.com/sarthak268/C3VQG-official}.} against state-of-the-art in VQG \cite{Krishna2019InformationMV, Wang2017AJM, Jain2017CreativityGD} using diverse quantitative metrics alongside highlighting the qualitative superiority of our approach.

\subsection{Dataset Features}

The VQA dataset \footnote{Dataset available at \href{https://visualqa.org/download.html}{https://visualqa.org/download.html}} \cite{VQA} consists of images along with corresponding questions and answers for each image. Additional information about the entire VQA dataset is presented in the supplementary.
Similar to works \cite{Krishna2019InformationMV, Wang2017AJM, Jain2017CreativityGD}, we have used the validation set as our test set due to lack of availability of ground-truth answers for the test set. 

\subsection{Evaluation Metrics}

We intend to evaluate our approach to compare it with prior work in VQG using a variety of language modeling metrics including \textit{BLEU}, \textit{METEOR} and \textit{CIDEr} \cite{Vedantam2014CIDErCI}. These metrics quantify the ability of the model to generate questions similar to ground-truth questions.

Additionally, we compute another quantitative metric ROUGE-L: a variant of ROUGE \cite{lin-2004-rouge}. This metric quantifies the similarity between generated and ground-truth questions using longest common sub-sequence. The advantage of using it is that it takes into account any structural association present at sentence level, capturing the longest \emph{n}-gram concurrently occurring in the sequence.

We also evaluate the performance of our model against the baselines using crowd-sourced metrics for testing the relevance of the generated question with respect to the ground-truth images and answer categories. For this, we conduct a user study among 5 crowd workers in which each is supposed to answer if the generated questions are consistent with respect to the given image and category.

In order to quantify the heterogeneity of generated questions, we additionally employ diversity metrics in our evaluation. For this, we compute \textit{strength} and \textit{inventiveness}.
While \textit{strength} is referred to as the percentage of unique generated question, \textit{inventiveness} is the ratio of unique generated questions unseen during training.

\subsection{Quantitative Results}
\label{quan_results}

\begin{figure}[h]
        \includegraphics[scale=0.3]{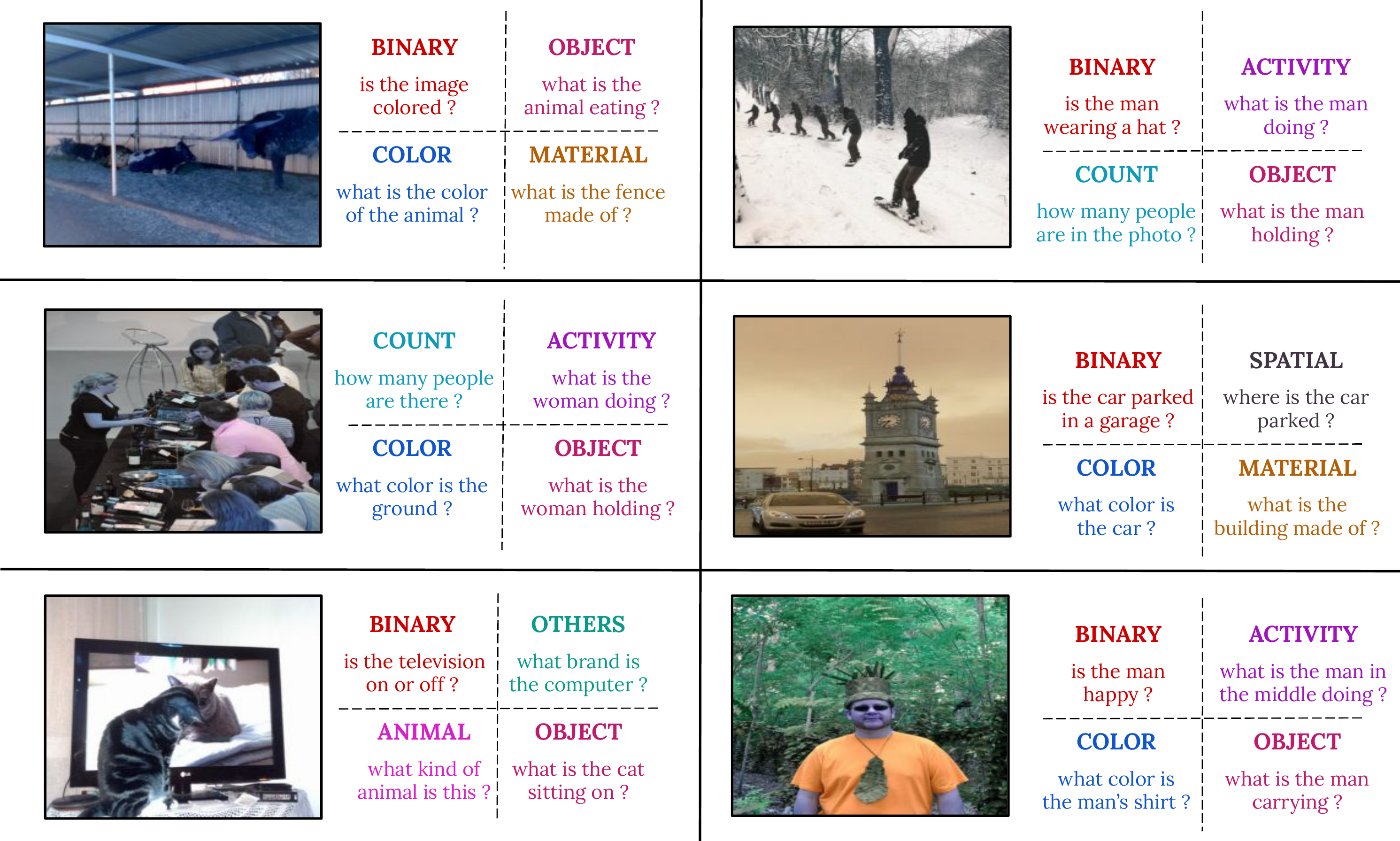}
        \caption{Questions generated for each image from multiple answer categories using ${\tt C3VQG}$ approach. }
        \label{fig:qualitative_all}
\end{figure}

\begin{table*}[t]
    \centering
    \begin{tabular}{cc|ccccccc}
    \hline
    \textbf{Supervision} & \textbf{Models} & \textbf{Bleu-1} & \textbf{Bleu-2} & \textbf{Bleu-3} & \textbf{Bleu-4} & \textbf{METEOR} & \textbf{CIDEr} & \textbf{ROUGE-L} \\
    \hline
    Supervised (\textit{w} A) 
    & IA2Q \cite{Wang2017AJM} & 32.43 & 15.49 & 9.24 & 6.23 & 11.21 & 36.22 & - \\
    
    & V-IA2Q \cite{Jain2017CreativityGD} & 36.91 & 17.79 & 10.21 & 6.25 & 12.39 & 36.39 & - \\
    
    & Krishna \textit{et al.} \cite{Krishna2019InformationMV} & 47.40 & 28.95 & 19.93 & 14.49 & 18.35 & 85.99 & 49.10 \\
    
    
    \hline
    \multirow{3}{*}{Weakly Supervised (\textit{w/o} A)} 
    
    & IC2Q \cite{Wang2017AJM} & 30.42 & 13.55 & 6.23 & 4.44 & 9.42 & 27.42 & - \\
    
    & V-IC2Q \cite{Jain2017CreativityGD} & 35.40 & \textbf{25.55} & 14.94 & \textbf{10.78} & 13.35 & 42.54 & - \\
    
    & Krishna \textit{et al.} \cite{Krishna2019InformationMV} \emph{w/o} A & 31.20 & 16.20 & 11.18 & 6.24 & 12.11 & 35.89 & 40.27 \\
    
    & \textit{I} & 38.44 & 19.83 & 12.02 & 7.69 & 13.27  & 45.19 & 40.90 \\
    & \textit{I + II} & 38.80 & 20.12 & 12.32 & 7.96 & 13.40 & 46.42 & 41.27 \\
    & \textit{I + CL} & 38.81 & 20.14 & 12.30 & 7.91 & 13.41 & 46.96 & 41.21\\
    & \textit{I + II + CL} & 38.94 & 20.30 & 12.47 & 8.10 & 13.47  & \textbf{47.32}  & 41.27 \\
    & \textit{I + II + Bayes} & 38.71 & 19.89 & 12.14 & 7.87 & 13.23 & 42.47 & 41.32 \\
    & \textit{I + CL + Bayes} & 38.64 & 20.06 & 12.28 & 7.95 & 13.32 & 45.83 & 41.16 \\
    & \textit{I + II + CL + Bayes} &  \textbf{41.87} & 22.11 & \textbf{14.96} & 10.04 & \textbf{13.60} & 46.87 & \textbf{42.34}\\
    \hline
\end{tabular}
\caption{Ablation study for different components of ${\tt C3VQG}$ using different language modeling quantitative metrics against other baselines in VQG. We compare our approach against previous state-of-the-arts in \textit{VQG}. } 
\label{quantitative}
\end{table*}

\begin{table}[h]
    \centering
    \begin{tabular}{c|cc|cc}
    \hline
    \textbf{Categories} & 
    \multicolumn{2}{c}{\textbf{Krishna \textit{et al.}} \cite{Krishna2019InformationMV}}
    
     & \multicolumn{2}{c}{\textbf{${\tt C3VQG}$}}
     \\
    
     & \textit{\textbf{S}} & \textit{\textbf{I}}
     & \textit{\textbf{S}} & \textit{\textbf{I}}  \\
     \hline
    
    ${\tt count}$ & 26.06 & 41.30 &  65.21 & 61.84 \\
    ${\tt binary}$ &  28.85 & 54.50 & 65.12 & 38.55 \\
    ${\tt object}$ & 24.19 & 43.20 & 65.58 & 58.85 \\
    ${\tt color}$ & 17.12 & 23.65 & 65.21 & 54.34 \\
    ${\tt attribute}$ & 46.10 & 52.03  & 64.59 & 63.02 \\
    ${\tt materials}$  & 45.75 & 40.72 & 64.87 & 63.48 \\
    ${\tt spatial}$ & 70.17 & 68.18 0 & 65.18 & 64.96 \\
    ${\tt food}$  & 33.37 & 31.19 & 65.20 & 62.21 \\
    ${\tt shape}$ & 45.81 & 55.65 & 66.01 & 65.98 \\
    ${\tt location}$  & 45.25 & 27.22 & 65.09 & 64.72 \\
    ${\tt predicate}$  & 36.20 & 31.29 & 65.67 & 65.67 \\
    ${\tt time}$  & 34.43 & 25.30 & 58.13 & 64.96 \\
    ${\tt activity}$ & 21.32 & 26.53 & 64.98 & 63.67 \\
    \hline
    Overall & 26.06 & 52.11 & \textbf{65.24} & \textbf{61.55} \\
    \hline
\end{tabular}
\caption{Quantitative evaluation of ${\tt C3VQG}$ against baselines using diversity metrics: Strength (\textit{\textbf{S}}) and Inventiveness (\textit{\textbf{I}}). Other comparisons present in the supplementary.}
\label{diversity_metrics}
\end{table}

\begin{table}[h]
    \centering
    \begin{tabular}{c|c|c}
    \hline
    \textbf{Model} & \multicolumn{2}{c}{\textbf{Relevance}} \\
    
     & \textbf{Image} & \textbf{Category} \\
    \hline
    
    V-IC2Q \cite{Jain2017CreativityGD} & 90.10 & 39.00 \\ 
    
    Krishna \textit{et al.} \cite{Krishna2019InformationMV} \emph{w/o} A & \textbf{98.10} & 42.70 \\
    
    \textit{${\tt C3VQG}$ \textit{w/o} Bayes, CL} & 98.00 & 58.40 \\
    
    \textit{${\tt C3VQG}$} & 97.80 & \textbf{60.50} \\
    \hline
\end{tabular}
\caption{Quantitative evaluation of ${\tt C3VQG}$ against other weakly supervised baselines using crowd-sourced metrics.}
\label{crowd_source_metrics}
\end{table}

In Table \ref{quantitative}, \textbf{I} and \textbf{II} depict step I and II respectively of our approach, \textbf{CL} depicts the imposed center loss on the combined latent space and \textbf{Bayes} represents an additional hyper-prior on the inverse variance of each latent dimension. Table \ref{quantitative} depicts that our approach beats state-of-the-art performance in VQG \cite{Krishna2019InformationMV} without answer supervision while training. The role of each component in the incremental build-up of our approach is clearly observable from the ablations reported. 
Additionally, it also shows the significance of cyclic consistency for generating category specific questions. Using multiple constraints on latent space reduces the performance slightly for \textit{Bleu-2} and \textit{Bleu-4}, but we observe significant increase in other language modelling metrics. We leave certain values for ROUGE-L blank in Table \ref{quantitative} as some prior works \cite{Wang2017AJM, Jain2017CreativityGD} did not employ it for their evaluation. 

The reported values in Table \ref{crowd_source_metrics} depict that our model outperforms baselines as a result of question-category consistency and the structure present in latent space. The incorporation of supplementary constraint on the congruence of answer category ensures the generated question's relevance to the category. Also, the squared \textit{L2} loss between the image encoding and encoding generated from the combined latent space assists relevance with respect to the image.

The superiority in the diversity of generated questions by our model as depicted in Table \ref{diversity_metrics} highlights that imposing a different prior on each dimension of the latent space enforces generation of a set of diversified questions from different answer categories. 

\subsection{Qualitative Results}
\label{qual_results}

\begin{figure}[h]
    \centering
    \includegraphics[scale=0.36]{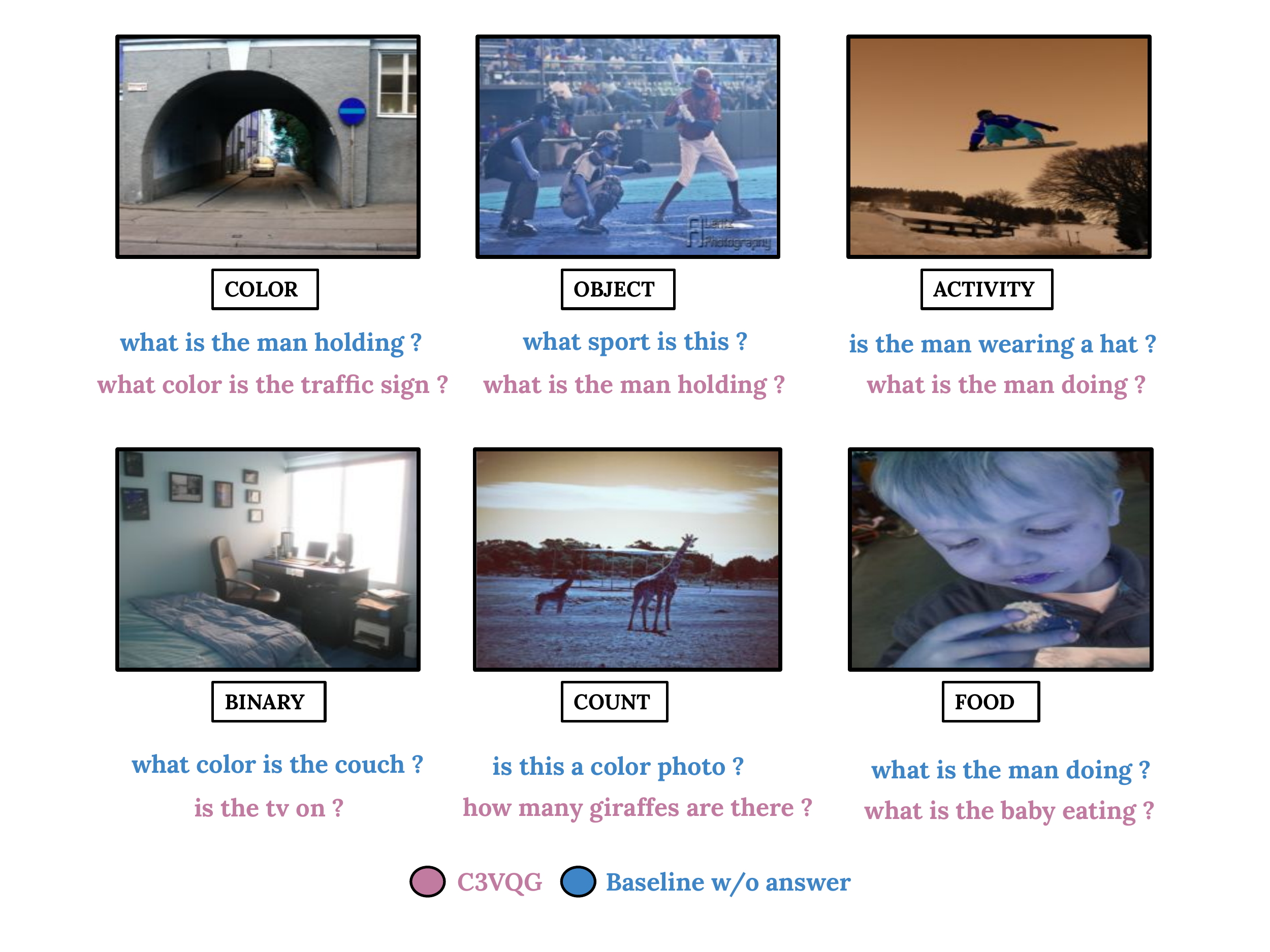}
    \caption{Qualitative results for ${\tt C3VQG}$ and Krishna \textit{et al.} \cite{Krishna2019InformationMV} without answers.}
    \label{fig:qualitative}
\end{figure}

We present a set of $4$ generated questions for a collection of images in Figure \ref{fig:qualitative_all}, demonstrating that our approach generates diverse image and category-consistent questions. Even for a particular category, the generations are not trivially replicated irrespective of the image. For \textit{e.g.}, as shown in the Figure \ref{fig:qualitative_all}, questions generated for the category ${\tt binary}$ are quite diverse for different images, thus, taking into consideration the context as well.

Additionally in Figure \ref{fig:qualitative}, we demonstrate cases in which the questions generated by our model belong to specified answer categories while the baseline approach in \cite{Krishna2019InformationMV} \textit{w/o} answer supervision fails to do so. For \textit{e.g.}, the top-left image of Figure \ref{fig:qualitative}, ${\tt C3VQG}$ is able to generate a question whose answer falls in the category of ${\tt color}$ whereas, for the question generated by the baseline approach \cite{Krishna2019InformationMV}, the answer category seems to be ${\tt object}$ instead of ${\tt color}$.

As demonstrated in the qualitative results, questions generated by \cite{Krishna2019InformationMV} are meaningful with respect to each image and are not generic, but they often lack correlation between categories and generated questions. We eradicate such inconsistencies of the generations with the provided categories by including cycle consistency and centre loss. 

\section{Conclusion}
\label{conclusion}

We present a novel category-consistent cyclic training approach ${\tt C3VQG}$ for visual question generation using structured latent space. Our approach generates category-specific comprehensive questions using visual features present in the image without using ground-truth answers. With weak supervision, our approach beats state-of-the-art in a variety of metrics. Qualitatively, our approach avoids generic question formation and generates category-consistent questions. While cyclic training helps in generating questions consistent with the answer category, the imposed latent structure ensures enhanced diversity of generations. This shows that effectively designing system configurations and imposing structured constraints can help frame better models even with minimum supervision.

As a further prospect to this work, we aim to analyze the efficacy of our approach in other \textit{QG} tasks such as conversational systems. We also intend to study the effect of such constraints on other multimodal tasks like image/text retrieval, image captioning, etc. for learning robust representations.

\section{Acknowledgements}

Rajiv Ratn Shah is partly supported by the Infosys Center for AI at IIIT Delhi.

  \bibliographystyle{ACM-Reference-Format}
  \bibliography{main}

\end{document}


\title{C3VQG: Category Consistent Cyclic Visual Question Generation \\ Supplementary Material}

\author{Shagun Uppal$^{1*}$, Anish Madan$^{1*}$, Sarthak Bhagat$^{1*}$, Yi Yu$^{2}$, Rajiv Ratn Shah$^{1}$}\thanks{$^*$Equal contribution. Ordered Randomly}
\affiliation{%
\institution{$^1$IIIT-Delhi, India; $^2$NII, Japan}
}
\email{{shagun16088, anish16223, sarthak16189, rajivratn}@iiitd.ac.in, yiyu@nii.ac.jp}

\renewcommand{\shortauthors}{Uppal et al.}

\maketitle

In this document, we begin by showing the final formulation of the loss functions to train our network, along with the detailed algorithm. We then derive the variational lower bound of the objective function of step 1 of our approach. We also  provide an illustration to depict the inference procedure of ${\tt C3VQG}$.
This is followed by listing the hyperparameters values and the details of VQA dataset used for training and evaluation of the ${\tt C3VQG}$ model. Lastly, we show some more qualitative results as well as the complete results of diversity-based performance of our model which showcases the strength of using our proposed hyper-prior on the latent space.

\section{Implementation Details}

Using the individual losses defined earlier, we aim to optimize a combined loss $\Lagr_{total}$ that is the weighted sum of individual loss terms.
Combining the individual losses defined earlier, we obtain the optimization objective as follows:

\begin{equation}
\begin{split}
    \min_{\mathbf{W}} \Lagr_{total} = \min_{\mathbf{W}} \Big[ \Lagr_{Q} + \lambda_{I}\Lagr_{I} + \lambda_{C}\Lagr_{C} + \lambda_{cons}\Lagr_{cons} \\ + \lambda_{center}\Lagr_{center} +  \lambda_{bayes}\Lagr_{bayes} \Big]
\end{split}
\end{equation}

where, $\mathbf{W}$ represents the combination of all learnable parameters in the complete model and $\lambda s$ are the hyperparameters depicting the weight of each loss in the combined objective.

\begin{algorithm}[h]
\SetAlgoLined
 \textbf{Input: } $dset$ containing $n$ training tuples of form $<i,q,C>$, multi-task loss weights for all individual losses: $\lambda s$, gradient descent learning rate $\alpha_{LR}$.
 
 \textbf{Output: } Optimal weights for all the individual components of the model $\mathbf{W}$.

 Initialize $\mathbf{W}$ with Kaiming initialization \cite{He2015DelvingDI}.
 
 \For{$epoch\gets1$ \KwTo $num\_epochs$}{
    \For{$k\gets1$ \KwTo $n$}{
        $i_k,q_k,C_k \leftarrow dset[k]$ \\
        Get $h^{i}_k$ and $h^{c}_k$ using Equation 3. \\
        Concatenate $h^{i}_k$ and $h^{c}_k$ to get $z_k$ using Equation 4. \\
        Use $z_k$ to predict $h^{i}_k$ and $h^{c}_k$ using networks $p^{i}$ and $p^{c}$ and compute $\Lagr_{I}$ and $\Lagr_{C}$ using Equation 7. \\
        Generate question $\hat{q}_{k,C_k}$ using Equation 5 and compute $\Lagr_{Q}$ using Equation 6. \\
        Predict category $C^{pred}_{k}$ from generated question using Equation 8 and compute $\Lagr_{cons}$ using Equation 9. \\
        Compute $\Lagr_{center}$ and $\Lagr_{bayes}$ using Equation 10 and 12 respectively. \\
        Find gradient of the total loss w.r.t. $\mathbf{W}$, i.e. $\nabla_{\mathbf{W}} \Lagr_{total}$. \\
        Take gradient descent step, $\mathbf{W} \leftarrow \mathbf{W} - \alpha_{LR} \nabla_{\mathbf{W}} \Lagr_{total}$. \\
    }   
 }
 \caption{Training Algorithm for ${\tt C3VQG}$ with all components.}
 \label{training_algo}
\end{algorithm}

For training our model using Algorithm \ref{training_algo}, we use gradient descent algorithm with Adam optimizer. 
We train the model for 15 epochs on a machine with single GeForce GTX 1080 GPU using the PyTorch framework.

\section{Derivation}
We re-iterate the objective function optimized in the main paper. The objective function is parameterized by $\phi$ which is optimized as follows:

\begin{align}
\label{eqn:mu_inf1}
    \max_\phi \quad I(q,z|i,C) + \lambda_1  I(i,z) + \lambda_2 I(C,z) \\  
    s.t \quad z \sim p_\phi(z|i,C) \\
    q \sim p_\phi(q|z)
\end{align}

where $\lambda_{1}$ and $\lambda_{2}$ are the weights for the mutual information terms.
The mutual information in Equation \ref{eqn:mu_inf1} is intractable as we do not know true values of the posteriors $p(z|i)$ and $p(z|C)$. So we instead try to minimize its variational lower bound (ELBO). In the equations below, $\mathbb{H}$ stands for entropy while $\mathbb{E}$ for expectation.

\begin{gather}\label{eq:info}
\begin{aligned}
\begin{aligned}
I(C,z) &= \mathbb{H}(C) - \mathbb{H}(C|z)
\\
 &= \mathbb{H}(C) + \mathbb{E}_{z \sim p(z,C)} [\mathbb{E}_{\hat{C}\sim p(C|z)}[\log p(\hat{C}|z)]]
 \\
 &= \mathbb{H}(C) + \mathbb{E}_{C \sim p(C)}[D_{KL}[p(\hat{C}|z) || p_\phi(\hat{C}|z)]] \\ 
 & \hspace{10mm} + \mathbb{E}_{\hat{C} \sim p(C|z)}[log p(\hat{C}|z)] \\
 & \geq \mathbb{H}(C) + \mathbb{E}_{C \sim p(C)}[\mathbb{E}_{\hat{C} \sim p(C|z)}[log p_\phi(\hat{C}|z)]] \\
\end{aligned}
\end{aligned}
\end{gather}

We similarly compute the expression for $I(i,z)$:
\begin{align}\label{eq:info1}
    I(i,z) \geq \mathbb{H}(i) + \mathbb{E}_{i \sim p(i)}[\mathbb{E}_{\hat{i} \sim p(i|z)}[log p_\phi(\hat{i}|z)]]
\end{align}

The expression for $I(q,z|i,C)$ then follows as:
\begin{align}\label{eq:info2}
    I(q,z|i,C) \geq \mathbb{H}(q) + \mathbb{E}_{q \sim p(q|i,C)}[\mathbb{E}_{\hat{q} \sim p(q|z,C,i)}[log p_\phi(\hat{q}|z,i,C)]] \\
    \text{where} \hspace{2mm} p(q|z,i,C) = p(q|z)p(z|i,C)
\end{align}

We substitute equations~\ref{eq:info}, \ref{eq:info1}, and \ref{eq:info2} in equation~\ref{eqn:mu_inf1}:
\begin{gather}\label{eq:final}
\begin{aligned}
\begin{aligned}
    & \max_\phi \hspace{5mm} \mathbb{E}_{p_\phi(q,i,C)}[log p_\phi(q|z,i,C) + \lambda_{1} log p_\phi(i|z) \\
    & \hspace{25mm} + \lambda_{2} log p_\phi(C|z)] \\
    & \hspace{10mm} \text{where} \hspace{3mm} p_\phi(q,i,C) = p_\phi(q|z) p_\phi(z|i,C) p_\phi(i,C) 
\end{aligned}
\end{aligned}
\end{gather}
 Hence, we can optimize the variational lower bound by maximizing the image and category reconstruction whilst also maximizing the MLE of question generation.




\section{Inference Framework}

\begin{figure*}[h]
        \centering
        \includegraphics[width=0.9\linewidth]{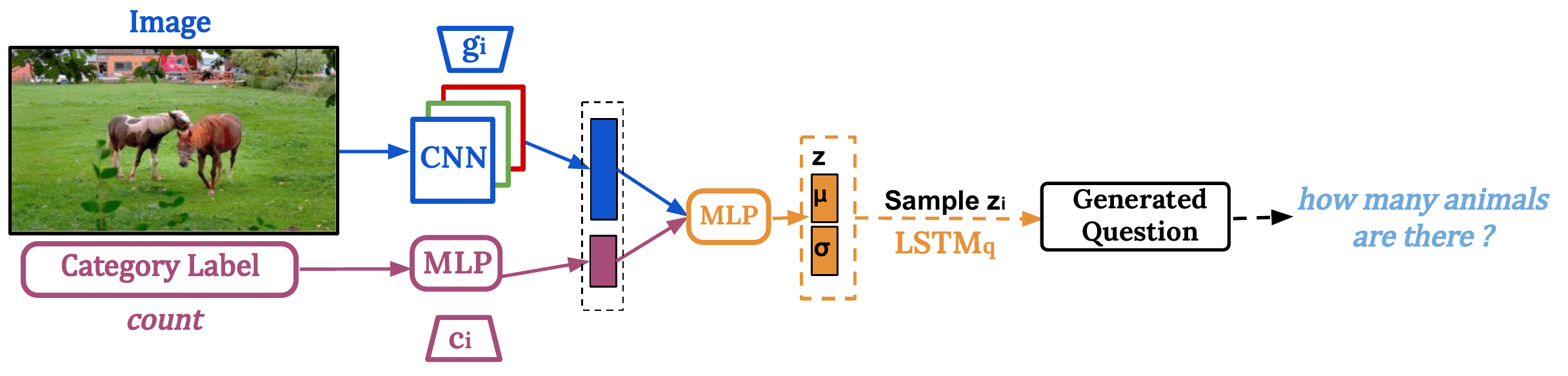}
        \caption{${\tt C3VQG}$ Inference Framework}
        \label{fig:inference}
\end{figure*}

We illustrate the inference flow using Figure \ref{fig:inference}. During inference, given an image conditioned on the category label, $z_{i}$ is sampled from the combined generative latent representation $z$ of the inputs learnt by the model. This representation is then passed through the temporal network $LSTM_{q}$, thereby, outputting the generated question. While the training of ${\tt C3VQG}$ requires images and their corresponding ground-truth questions from different answer categories, the inference only requires the images with answer categories of the questions to be generated.

\section{Hyperparameters}

We present all a list of all the hyperparameter values used in training the ${\tt C3VQG}$ model.

\begin{table}[H]
    \centering
    \begin{tabular}{c|c|c}
    \hline
    \textbf{Hyperparameter} & \textbf{Symbol} & \textbf{Value} \\
    \hline
    Image Recon. Weight & $\lambda_{I}$ & 1.0 \\
    Category Recon. Weight & $\lambda_{C}$ & 2.0 \\
    Question Recon. Weight & $\lambda_{Q}$ & 3.0 \\
    Category Consistency Weight & $\lambda_{cons}$ & 2.0 \\
    Center Loss Weight & $\lambda_{center}$ & 3.0 \\
    Hyper-prior KL-Divergence Weight & $\lambda_{bayes}$ & 3.0 \\
    Hyper-prior Regularisation Weight & $\lambda_{reg}$ & 2.0 \\
    Dimension of combined latent space & $d$ & 64\\
    Learning Rate & $\alpha_{LR}$ & 1e-3 \\
    \hline
\end{tabular}
\caption{Hyperparameters values used for training ${\tt C3VQG}$.}
\label{hyperparams}
\end{table}

\section{Dataset Details}

We list the details about the VQA dataset \cite{VQA} used for the training and evaluation of ${\tt C3VQG}$ against the state-of-the-art \cite{Krishna2019InformationMV, Wang2017AJM, Jain2017CreativityGD} in VQG.

\begin{table}[H]
    \centering
    \begin{tabular}{c|c|c}
    \hline
    \textbf{Data Type} & \textbf{Training} & \textbf{Validation} \\
    \hline
    VQA Annotations (answers) & 4,437,570 & 2,143,540 \\
    VQA Input Questions & 443,757 & 214,354 \\
    VQA Input Images & 82,783 & 40,504 \\
    \hline
\end{tabular}
\caption{Dataset details for the VQA dataset.}
\label{hyperparams}
\end{table}

Krishna \textit{et al.} \cite{Krishna2019InformationMV} annotates the answers with a set of 15 categories and labels their top 500 answers. This makes up 82\% of the entire VQA dataset consisting of 367K training and validation examples. We use a 80:20 training-validation split for our experiments.

\section{Qualitative Results}

\begin{figure*}[h]
        \includegraphics[scale=0.35]{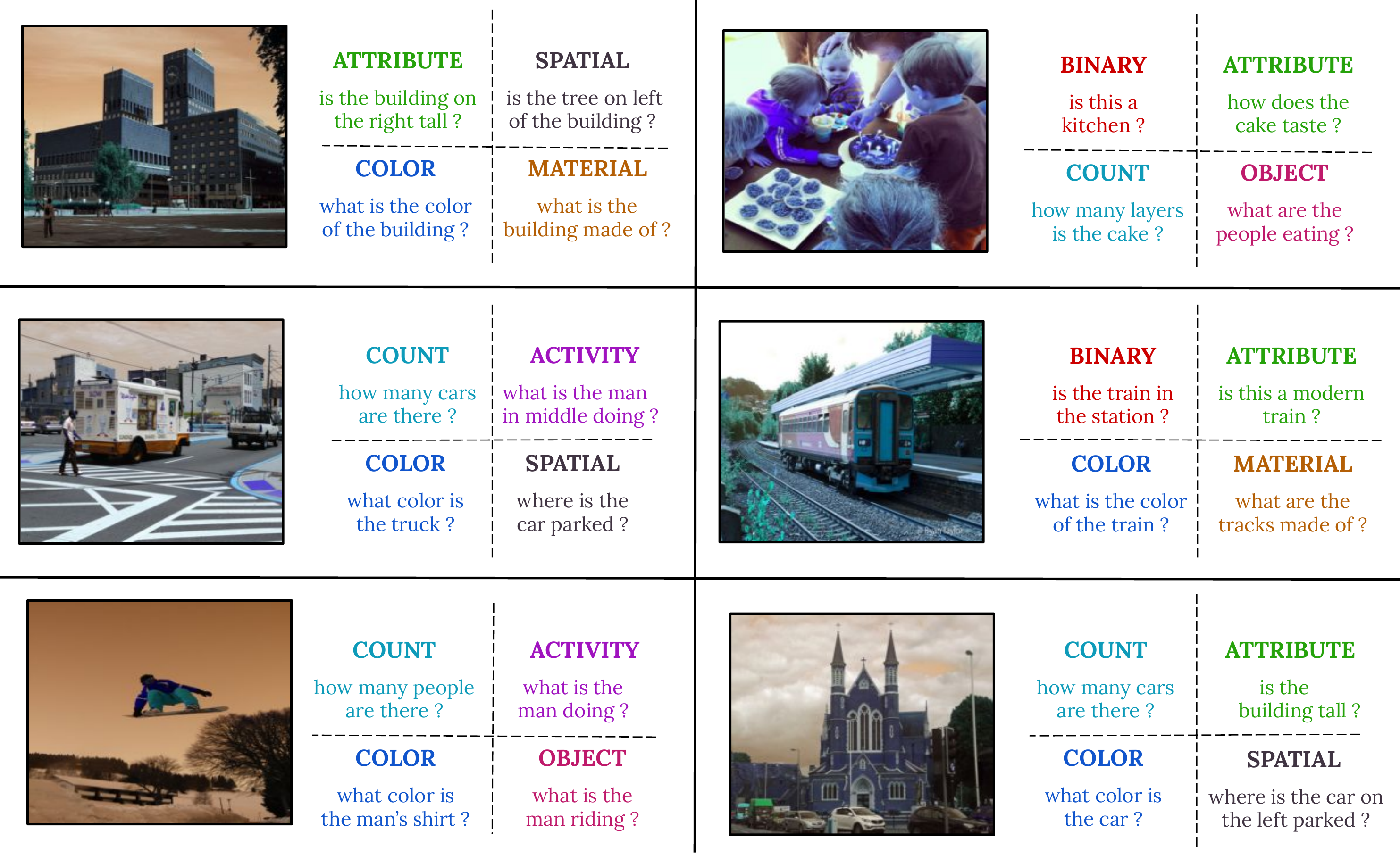}
        \caption{Questions generated for each image from multiple answer categories using ${\tt C3VQG}$ approach. }
        \label{fig:qualitative_all}
\end{figure*}

We depict additional qualitative results wherein we mention four questions (from different answer categories) generated using our approach on a collection of images from the validation set. As depicted in Figure \ref{fig:qualitative_all}, the questions generated are non-generic and category-specific for each of the following examples.

\section{Diversity-based Performance Evaluation}

Here, we depict additional comparisons in terms of the diversity-based metrics.
As depicted in Table \ref{diversity_metrics}, the performance in terms of the diversity of generated questions achieved by our approach with all components beats the state-of-art in VQG even without the requirement of additional answer supervision. The difference in the \textit{strength} and \textit{inventivenes} values with and without the latent hyper-prior suggests that capturing decorrelated features in each latent dimension enables our model to generate non-generic questions from a divergent pool of categories.

\begin{table*}[h]
    \centering
    \begin{tabular}{c|cc|cc|cc|cc}
    \hline
    \textbf{Categories} & \multicolumn{2}{c}{\textbf{V-IC2Q} \cite{Jain2017CreativityGD}} &
    
    \multicolumn{2}{c}{\textbf{Krishna \textit{et al.}} \cite{Krishna2019InformationMV}}
    
     & \multicolumn{2}{c}{\textbf{${\tt C3VQG}$ w/o Bayes}}
     & \multicolumn{2}{c}{\textbf{${\tt C3VQG}$}}
     \\
    
     & \textbf{Strength} & \textbf{Inventiveness}
     & \textbf{Strength} & \textbf{Inventiveness}
     & \textbf{Strength} & \textbf{Inventiveness} & \textbf{Strength} & \textbf{Inventiveness}  \\
     \hline
    
    ${\tt count}$ & 15.77 & 30.91 & 26.06 & 41.30 & 58.33 & 55.20 &  65.21 & 61.84 \\
    ${\tt binary}$ & 18.15 & 41.95 &  28.85 & 54.50 & 58.39 & 36.32 & 65.12 & 38.55 \\
    ${\tt object}$ & 11.27 & 34.84 & 24.19 & 43.20 & 57.77 & 51.51 & 65.58 & 58.85 \\
    ${\tt color}$ & 4.03 & 13.03 & 17.12 & 23.65 & 58.38 & 48.97 & 65.21 & 54.34 \\
    ${\tt attribute}$ & 37.76 & 41.09 & 46.10 & 52.03 & 60.05 & 58.38 & 64.59 & 63.02 \\
    ${\tt materials}$ & 36.13 & 31.13 & 45.75 & 40.72 & 57.93 & 56.79 & 64.87 & 63.48 \\
    ${\tt spatial}$ & 61.12 & 62.54 & 70.17 & 68.18 & 57.90 & 57.80 & 65.18 & 64.96 \\
    ${\tt food}$ & 21.81 & 20.38 & 33.37 & 31.19 & 58.49 & 55.42 & 65.20 & 62.21 \\
    ${\tt shape}$ & 35.51 & 44.03 & 45.81 & 55.65 & 58.85 & 58.75 & 66.01 & 65.98 \\
    ${\tt location}$ & 34.68 & 18.11 & 45.25 & 27.22 & 58.39 & 58.10 & 65.09 & 64.72 \\
    ${\tt predicate}$ & 22.58 & 17.38 & 36.20 & 31.29 & 57.05 & 57.05 & 65.67 & 65.67 \\
    ${\tt time}$ & 25.58 & 15.51 & 34.43 & 25.30 & 58.13 & 58.10 & 65.00 & 64.96 \\
    ${\tt activity}$ & 7.45 & 13.23 & 21.32 & 26.53 & 58.00 & 56.78 & 64.98 & 63.67 \\
    \hline
    Overall & 12.97 & 38.32 & 26.06 & 52.11 & 58.23 & 54.99 & \textbf{65.24} & \textbf{61.55} \\
    \hline
    
    \hline
\end{tabular}
\caption{Quantitative evaluation of ${\tt C3VQG}$ against other baselines using diversity-based metrics.}
\label{diversity_metrics}
\end{table*}

\bibliographystyle{ACM-Reference-Format}
\bibliography{supp}